\pdfoutput=1

\documentclass[11pt]{article}
\usepackage{soul}
\usepackage{tablefootnote}
\usepackage{parskip}

\usepackage{ACL2023}
\usepackage{float}

\usepackage{times}
\usepackage{latexsym}

\usepackage[T1]{fontenc}

\usepackage[utf8]{inputenc}

\usepackage{microtype}

\usepackage{inconsolata}
\counterwithout{figure}{section}
\counterwithout{figure}{subsection}

\usepackage{booktabs}
\usepackage{makecell}
\usepackage{longtable}
\usepackage{amsmath}
\usepackage{amsfonts}
\usepackage{bbm}
\usepackage{cleveref}

\title{ACORD: An Expert-Annotated Dataset for Legal Contract Clause Retrieval}

\author{Steven H. Wang$^{1}$, Maksim Zubkov$^{2}$, Kexin Fan$^{3}$, Sarah Harrell$^{4}$, Yuyang Sun$^{5}$ \\\ \bf
Wei Chen$^{6}$, Andreas Plesner$^{1}$, Roger Wattenhofer$^{1}$ \\
$^1$ETH Zurich $^2$Independent Researcher $^3$New York University \\
$^4$University of Washington $^5$Yale University $^6$The Atticus Project \\
\texttt{wang.steven.h@gmail.com}\hspace{0.2cm} \texttt{wei@atticusprojectai.org}\hspace{0.1cm} \texttt{aplesner@ethz.ch}}

\begin{document}\maketitle
\begin{abstract}
Contract clause retrieval is foundational to contract drafting because lawyers rarely draft contracts from scratch; instead, they locate and revise the most relevant precedent clauses. We introduce the Atticus Clause Retrieval Dataset (ACORD), the first expert-annotated benchmark specifically designed for contract clause retrieval to support contract drafting tasks. ACORD focuses on complex contract clauses such as Limitation of Liability, Indemnification, Change of Control, and Most Favored Nation. It includes 114 queries and over 126,000 query-clause pairs, each ranked on a scale from 1 to 5 stars. The task is to find the most relevant precedent clauses to a query. The bi-encoder retriever paired with pointwise LLMs re-rankers shows promising results. However, substantial improvements are still needed to manage the complex legal work typically undertaken by lawyers effectively. As the first expert-annotated benchmark for contract clause retrieval, ACORD can serve as a valuable IR benchmark for the NLP community. 
\end{abstract}

\section{Introduction}
\begin{figure}[ht]
\centering
\scriptsize
    \includegraphics[width=\linewidth]{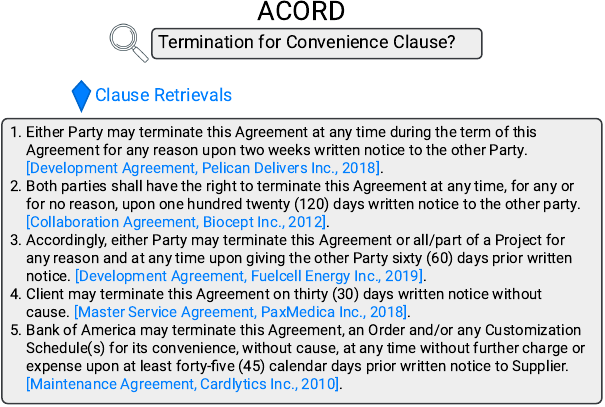}
    \vspace{-.2cm}
    \caption{
    A showcase of the clause retrieval process integral to the ACORD dataset. The input is a query, ``Termination for Convenience,'' and the output is a ranked list of the most relevant clauses from various agreements. The clauses are accompanied by their sources, thus providing critical context for legal professionals.
    }
    \label{fig: ACORD search}
\end{figure}

Contracts are the backbone of modern business, with millions created daily. A 2023 ALM and Bloomberg Law survey found that 43\% of corporate counsels spend at least half their time on tasks like drafting, editing, and negotiating contracts (\citet{BloombergLaw2023,pery_contract_2019}). Clause retrieval is critical to these tasks because lawyers rarely write contracts from scratch; they rely on finding and adapting high-quality precedent clauses, especially when drafting complex clauses. Mistakes in wording or missing key ideas can lead to disputes, liabilities, or invalid clauses.

Research shows that Large Language Models (LLMs) can effectively analyze legal contracts and identify issues within them \citep{hoffman_generative_2023, martin_better_2024,choiLawyeringAgeArtificial2024,schwarczAIPoweredLawyeringAI2025}. However, LLMs struggle to draft contracts independently. \Cref{tab: redlining comparison} compares an LLM-generated clause with two top-rated precedent clauses from ACORD. Lawyer edits, shown with strike-through and underlined text, highlight two main issues: (1) the LLM-generated language creates conflicts between clauses, and (2) it introduces language not typically found in standard contracts.

\begin{table*}[t]
    \centering
    \renewcommand{\arraystretch}{1.5}
    \setlength{\tabcolsep}{5pt}
    \scriptsize
    \begin{tabular}{p{0.46\textwidth}|p{0.23\textwidth}p{0.23\textwidth}}
        \toprule
        \textbf{Prompt: draft a contract clause where liability cap is based on purchase price} 
        & \multicolumn{2}{c}{\textbf{Query: liability cap is based on purchase price}} \\
        \midrule
        \textbf{LLM-Generated Result}& \textbf{ACORD 5-Star Clause} & \textbf{ACORD 4-Star Clause} \\
        \midrule
        \st{1.} \textcolor{red}{\underline{[Section Number]}} Cap on Liability$^a$

        \st{Notwithstanding any provision to the contrary in this Agreement, }
        \textcolor{red}{\underline{Except as set forth}} \textcolor{red}{\underline{in Section [Number] below,}}$^b$ \st{T}\textcolor{red}{t}he total aggregate liability of [Seller's Name] to [Buyer's Name], whether in contract, tort (including negligence), breach of statutory duty, or otherwise, arising under or in connection with this Agreement shall not exceed the total \st{purchase price}\textcolor{red}{amount} paid \textcolor{red}{\underline{and payable}} by [Buyer's Name] to [Seller's Name] under this Agreement (the ``Purchase Price''). $^c$

        \st{2.} \textcolor{red}{\underline{[Section Number].}} Exclusions from Liability Cap

        The limitation of liability set forth in \st{this clause} \textcolor{red}{\underline{Section [Number]}} shall not apply to:
        
        a. Liability resulting from gross negligence, willful misconduct, or fraud by [Seller's Name];
        
        b. Any breach of confidentiality obligations as specified in this Agreement;
        
        c. Any liability which cannot be excluded or limited by applicable law.$^d$

        \st{3.} \textcolor{red}{\underline{Section [Number]}} No Indirect or Consequential Damages
        
        In no event shall [Seller's Name] be liable to [Buyer's Name] for any indirect, incidental, special, consequential, or punitive damages, including but not limited to loss of profits or revenue, loss of use, or loss of data, even if advised of the possibility of such damages.

        \st{4. Application of Limitations
        
        The parties agree that the limitations of liability set forth in this clause are fundamental elements of the basis of the bargain between [Seller's Name] and [Buyer's Name], and the Purchase Price reflects these limitations.}$^e$
        & 5. Sponsor's Maximum Obligation; Indemnification. Racing represents to the Sponsor that the Sponsor's aggregate obligation hereunder will not exceed the amount of the sponsorship fee set forth in Section 3 hereof (or such lesser amount as is payable by the Sponsor in the event that this Agreement is terminated pursuant to Section 14 hereof), plus, if applicable, collection costs that may be reasonably incurred by Racing in a legal proceeding to collect all or any part thereof (the ``Maximum Obligation''). Racing agrees to indemnify the Sponsor and its officers, directors, agents and employees and to hold them harmless from any loss, claim, cost, damage or liability in excess of the Maximum Obligation which (i) the Sponsor shall incur as a result of this Agreement, or (ii) arises from any failure by Racing to perform any of its obligations hereunder.
        & 12. Limitation of Liability
        
        12.1 IN NO EVENT SHALL EITHER PARTY BE LIABLE TO THE OTHER PARTY FOR ANY INCIDENTAL, CONSEQUENTIAL, INDIRECT, SPECIAL, OR PUNITIVE DAMAGES (INCLUDING, BUT NOT LIMITED TO, LOST PROFITS, BUSINESS INTERRUPTION, LOSS OF BUSINESS INFORMATION OR OTHER PECUNIARY LOSS) REGARDLESS OF WHETHER SUCH LIABILITY IS BASED ON BREACH OF CONTRACT, TORT (INCLUDING NEGLIGENCE), STRICT LIABILITY, BREACH OF WARRANTIES, FAILURE OF ESSENTIAL PURPOSE OR OTHERWISE AND EVEN IF ADVISED OF THE POSSIBILITY OF SUCH DAMAGES.
        
        12.2 Except with regard to a breach of confidentiality, a party's indemnification obligations hereunder, or infringement of intellectual property rights, either party's total liability to the other party under this Agreement shall be limited to the amounts paid or payable by the Reseller to Todos during the twelve-month period preceding the interposition of the claim.\\
    \bottomrule
    \end{tabular} \\
    \parbox{\linewidth}{%
    \raggedright
    \scriptsize \textsuperscript{a} Attorney Note: change ``1.'' to a placeholder because a Cap on Liability clause would never be in Section 1.\\
    \scriptsize \textsuperscript{b}
    Attorney Note: the reference to ``Notwithstanding any provision to the contrary in this Agreement'' is wrong because this subsection should NOT take precedence over the ``Exclusions from Liability Cap'' below. Therefore, it should be replaced with ``Except as set forth in Section [Number] below.''\\
    \scriptsize \textsuperscript{c} Attorney Note: liability cap should not be at purchase price paid in the event the buyer hasn’t paid in full or fails to pay timely. ACORD 5-star and 4-star clauses fix this defect by capping the liability at the amount ``paid or payable'' or referring to the amount in the payment section of the contract.\\
    \scriptsize \textsuperscript{d} Attorney’s Note: This is not wrong; however, Exceptions to Liability Cap are often included in the same sentence as the liability cap instead of being in a stand-alone subsection.\\
    \scriptsize \textsuperscript{e} Attorney Note: Section 4 is not commonly seen in commercial contracts. It is unclear what the intent or purpose of this clause is.
    }
    \caption{
    Comparison between an LLM-generated liability cap clause and two expert-rated clauses from ACORD. Attorney annotations (strike-through and underlined text) highlight deficiencies in the LLM output, demonstrating why retrieval of high-quality precedent clauses is essential for effective contract drafting.
    }
    \label{tab: redlining comparison}

\end{table*}

Retrieval-Augmented Generation (RAG), which combines LLMs' text generation with a retrieval system, mimics how lawyers improve drafting by first searching for relevant precedent clauses to use as the base. This approach has been shown to reduce hallucinations and enhance performance \citep{lewis_retrieval-augmented_2021, niu_ragtruth_2024, magesh_hallucination-free_2024}. We seek to evaluate how well models retrieve relevant clauses to ensure RAG can work effectively.

Clause retrieval poses several challenges. Contracts are multi-layered, with sections and subsections that can span pages, often broken into paragraphs outlining obligations (what is required) and exceptions (what is not required). Cross-references within clauses add further complexity, such as defined terms (e.g., ``as defined in Section 4'') or exceptions (e.g., ``except as set forth in Sections 9 and 11(a)'' or ``notwithstanding anything to the contrary''). These references often point to sections located pages apart, yet they are crucial for understanding a clause’s meaning. Additionally, whether one clause should rank higher than the others can be subjective and depends on the user’s individual experience, the industry they are in, and their unique needs, among other things. This subjectivity is reflected in the annotators' disagreement rate of 21\%.\footnote{The fraction of ratings where the annotators significantly disagreed on the score or where they disagreed on the relevance. The value is common for hard datasets in the area. Under Labeling Process in \Cref{sec: acord}, we explain how we mitigate this issue through annotator selections, guideline design, and expert-panel review.}

Datasets with expert-annotated, domain-specific benchmarks are essential for improving LLMs in clause retrieval and contract drafting. However, such datasets are rare due to the high cost of expert annotations and the confidentiality of proprietary contracts. Many existing legal contract benchmarks consist of multiple-choice Q\&A formats, often derived from bar exam questions, and do not reflect the work lawyers perform, with none focusing specifically on contract drafting. To address this, we introduce ACORD, the Atticus Clause Retrieval Dataset, the first expert-annotated retrieval benchmark explicitly designed for contract drafting.

ACORD addresses the challenge of retrieving complex and heavily negotiated contract clauses, such as Limitation of Liability, Indemnification, Most Favored Nation, and Termination for Convenience. These clauses are central to contract drafting and require careful negotiation and precise language. The task is to retrieve the most relevant precedent clauses for a given query, where the output is a ranked list of top-rated clauses from a large corpus. \Cref{fig: ACORD search} illustrates this retrieval process.
The dataset was developed collaboratively by seasoned lawyers, student annotators, and machine learning researchers. It comprises 114 unique queries and over 126,000 query-clause pairs, each annotated with a 1-to-5-point relevance score by experts. This rigorous annotation ensures that ACORD is a robust and reliable benchmark for evaluating retrieval models in legal contexts. The estimated annotation cost of ACORD would be over US \$1 million when using the prevailing rates of \$550 per hour for attorneys and \$150 for non-attorneys.


ACORD provides a new framework for assessing machine learning models' ability to retrieve information critical for contract drafting tasks. We hope that ACORD enables researchers and ML practitioners to evaluate search systems properly on a nuanced legal search task and to make progress toward tackling real-world challenges in legal drafting.
The dataset and software will be made publicly available on GitHub.

\section{Related Work}

Despite being one of the most common legal tasks \citep{pery_contract_2019, BloombergLaw2023}, contract drafting and clause retrieval have been under-studied in NLP literature due to the lack of domain-specific benchmark datasets. 

\subsection{Information Retrieval}

Information retrieval (IR) is the process of locating relevant documents in response to a user query, as seen in web search tools such as Google and Bing. Historically, lexical-based retrieval has been foundational in IR \citep{hambarde_information_2023}, with BM25 being one of the most popular ranking functions. BM25 scores document relevance based on term frequency and inverse document frequency while mitigating the impact of prevalent words \citep{robertson_probabilistic_2009}. However, lexical-based retrieval methods face the ``lexical gap'', where reliance on exact word matches often fails to capture relevant semantic connections such as synonymy and word-order-dependent relationships \citep{thakur_beir_2021}. Recent approaches have moved toward neural IR, designed to capture semantic connections beyond simple lexical matches, to mitigate this. These IR models fall into two primary categories:

\paragraph{Retriever Models} Queries and documents are mapped into a vector space by the models, allowing pre-computed document representations to be indexed. For example, dense retrievers based on Transformer models \citep{vaswani} have shown robust performance in many tasks such as citation prediction, argument retrieval, and question-answering \citep{thakur_beir_2021, muennighoff_mteb_2023}. Another example is sparse embeddings that learned token-level contextualized representations using Transformers that still could be efficiently indexed using an efficient inverse index \citep{zhao_sparta_2020, formal_splade_2021}.

\paragraph{Neural Reranking Models} Initial retrieval results (often from e.g. a BM25 model) are enhanced by the models by reordering the returned documents. Leveraging cross-attention mechanisms in models such as BERT has significantly boosted reranking performance \citep{devlin_bert_2019}. ColBERT (Contextualized late interaction over BERT) refines this approach by creating token-level embeddings for queries and documents, using a maximum-similarity mechanism to identify relevant documents. However, this comes with higher computational costs \citep{khattab_colbert_2020}. Recent research highlights that instruction-tuned LLMs can surpass traditional supervised cross-encoders in zero-shot passage reranking tasks \citep{sun_is_2023}. In particular, list-, point-, or pairwise reranking have received attention \citep{tang_found_2024, zhuang_beyond_2024, qin_large_2024}.


\subsection{Large Language Models (LLMs) and Retrieval-Augmented Generation (RAG)}

Despite their advanced text generation capabilities, LLMs have not been widely adopted in contract drafting due to the risk of hallucination. Existing works \citep{wang_legal_2024, dahl_large_2024} show that the generations of LLMs often can be unreliable, untrustworthy, and risky. Many recent studies focus on developing and enhancing various RAG techniques to reduce hallucination and improve model performance in text generation \citep{niu_ragtruth_2024, magesh_hallucination-free_2024,schwarczAIPoweredLawyeringAI2025}. Given the absence of an information retrieval-specific dataset tailored to this task, the open question is whether these methodologies can effectively retrieve precedent clauses for contract drafting purposes.

\subsection{Case Law and Statute Retrieval}

There are several case law retrieval datasets, including LePARD, CLERC, COLLIE, ECtHR-PCR \citep{li_lepard_2022, hou_clerc_2024, rabelo_coliee_2021, tyss_ecthr-pcr_2024}. However, the queries in these datasets are not constructed by experts. Instead, they are synthetically created by taking blocks of text from case law with the citations redacted. The document referred to in the redacted citation serves as the corresponding relevant document for the query-document pair. This methodology cannot be used to create a contract IR dataset because contract clauses do not inherently include references or citations that can be used to infer query-document pairs. Legal experts need to generate queries and ranked clauses. 

Despite the relative abundance of case law IR datasets, case law retrieval remains an active and challenging area of research. Recent benchmarks report low baseline performance even when using forgiving metrics such as recall@1000 (CLERC) \citep{hou_clerc_2024} or constructing queries only from text containing direct quotations from the relevant citations (LePARD) \citep{li_lepard_2022}.

\subsection{Legal Contract Datasets}

Several datasets have been published to facilitate AI research on legal contracts, but none address specific IR needs. 


Prior datasets focus on clause extraction rather than retrieval. \citet{chalkidis_extracting_2017} \& \citet{leivaditi_benchmark_2020} provide datasets for extracting basic contract and lease information, not legal clauses, making them less useful for contract drafting. \citealp{hendrycks_cuad_2021} introduces CUAD, a large-scale expert-annotated dataset with over 2,000 clauses across 41 categories from 510 commercial contracts. While useful for drafting, CUAD is designed for clause classification and extraction and therefore lacks ranking, an essential component of IR datasets. MAUD \citep{wang_maud_2023}, ContractNLI \citep{koreeda_contractnli_2021}, and LegalBench \citep{guha_legalbench_2023} focus on reading comprehension and reasoning, but would need extensive expert annotations for IR research. BigLaw Bench (Core, Workflow, and Retrieval) \citep{noauthor_harveyaibiglaw-bench_2024} targets day-to-day legal tasks but is not specific to contract drafting and has an order of magnitude fewer annotations than ACORD.

\citet{aggarwalClauseRecClauseRecommendation2021,joshiInvestigatingStrategiesClause2022,lamApplyingLargeLanguage2023,alonsoAutomatingLegalContracts2024,kasundraAdaptingOpenSourceLLMs2024} have amongst others explored generating contract clauses by learning from data or applying constraints like logic rules or similar contract context. However, they generally rely on small datasets with simple queries and clauses. What is missing is a large-scale expert-rated benchmark that reflects how lawyers search for complex clauses, such as limitation of liabilities. 
ACORD fills this gap by providing the first large-scale benchmark for clause retrieval with queries crafted by experienced lawyers to fit their real needs. 

\section{ACORD: A Contract Clause Retrieval Dataset}\label{sec: acord}

\begin{table}[t]
    \renewcommand{\arraystretch}{1.5}
    \small
    \centering
    \begin{tabular}{l|m{4cm}}
        \toprule
        \textbf{Clause Category} & \textbf{Example Query} \\
        \midrule
        Limitation of Liability & a party's liability for fraud, negligence, personal injury or tort subject to a cap \\
        
        Indemnification & indemnification of first party claims \\
        
        Affirmative Covenants & insurance clause \\
        
        Restrictive Covenants & no solicit of customers \\
        
        Term \& Termination & termination for convenience \\
        
        Governing Law & clause with multiple governing laws \\
        \bottomrule
    \end{tabular}
    \caption{
    A subset of the 114 unique queries created by legal experts to address diverse contract drafting scenarios. Each query corresponds to a specific clause category, such as ``Limitation of Liability.'' 
    The full list of the clause categories and queries is in \Cref{appendix: data list and statistics}.
    }
    \label{tab: sample queries}
\end{table}




\paragraph{Clause Retrieval.} Clause retrieval involves identifying the most relevant precedent clauses within contracts based on a given query. This process ensures that lawyers can efficiently reference and adapt precedent clauses to meet specific drafting needs.

\paragraph{Task Definition.} ACORD frames the retrieval of clauses as an ad hoc information retrieval task. Given an attorney-written query and a corpus of clauses, search systems output a list of clauses by order of predicted relevance to the query. Performance is a function of the ordered lists (sometimes called a ``run'') and the attorney's ground-truth ratings. We measure performance with established metrics, such as the standard NDCG score and a task-specific normalized precision score.

\paragraph{Queries and Clauses.} The dataset includes 114 unique queries written by legal experts to address diverse drafting requirements. Each query targets one or more legal concepts across 9 clause categories. 
Sample queries are shown in \Cref{tab: sample queries}, with a complete list provided in \Cref{appendix: data list and statistics}.

\paragraph{Query-Clause Pair Score.} Each query-clause pair is assigned a 1-5 point relevance score. Clauses rated 3-5 stars are relevant, 2-star clauses are not relevant but helpful (from the same category, e.g., indemnification), and 1-star clauses are irrelevant. A sample-scored query-clause pair is shown in \Cref{tab: sample query-clause score}, and the annotation rubrics are in \Cref{tab: annotation rubrics}.

For each query, annotators aim to find 10 relevant clauses (3-5 stars) and 20 2-star clauses. Some queries have fewer than 10 relevant clauses, but all include at least 5. To avoid false negatives, 1-star clauses are selected from expert-annotated clauses in CUAD \citep{hendrycks_cuad_2021}, and 2-5 star Limitation of Liability and Indemnification clauses serve as 1-star for other categories. Though 1-star clauses are not individually rated, they are irrelevant due to being from distinct categories. \Cref{tab: clause statistics} shows the statistics of rated clauses, and \Cref{tab: ACORD queries and clauses counts} lists the number of clauses by rating for each category.

Since most queries have fewer than five 5-star or 4-star clauses, x-star precision@5 scores are normalized to a 0-1 scale to adjust for the limited availability of high-rated clauses, as explained in \Cref{appendix: performance metrics}. When calculating the precision, the scores are rounded: 4.666 rounds up to 5, 4.333 rounds down to 4, etc.
    
\paragraph{Contract Corpus and Clause Corpus.} About 400 of such contracts are from CUAD ~\citep{hendrycks_cuad_2021}, which was sourced from the Electronic Data Gathering, Analysis, and Retrieval (EDGAR) system maintained by the U.S. Securities and Exchange Commission (SEC).
50 contracts are Terms of Services, ToS, published online by selected Fortune 500 companies. We call these EDGAR contracts and online ToS the ``Contract Corpus.'' Annotators extracted from the Contract Corpus all clauses responsive to the 9 categories, which we call the ``Clause Corpus.'' Annotators then extracted clauses from the Clause Corpus responsive to each of the 114 queries. When extracting a clause responsive to a query, annotators include the entire subsection or section to ensure comprehensive context and understanding. As shown in \Cref{fig: word count distribution}, clauses in ACORD vary in length, with over half of the 3,000 clauses rated 2- through 5-stars having over 100 words. To assess the models’ retrieval performance, ACORD simplifies the real-life task of clause retrieval, which would require the models first to extract the Clause Corpus from the Contract Corpus. The benchmark included in this paper uses the Clause Corpus only.

\paragraph{Data Splits} We split the Clause Corpus into train, validation, and test sets. We form these splits at the query level, randomly allocating 45\%, 5\%, and 50\% of the queries to the train, validation, and test splits, respectively, while ensuring that at least one query from each category is represented in the test set.

\paragraph{Data Statistics} ACORD contains 114 unique queries across the 9 clause categories. It has over 126,000 query-clause pairs, each rated with a 1-5 score. See \Cref{appendix: data list and statistics} for more details of the queries and expert-rated clauses. 

\paragraph{No False Negatives}  Information retrieval datasets often contain false negatives due to an inability to annotate the entire dataset; however, every clause in ACORD is annotated to avoid false negatives. 

\paragraph{Labeling Process}\label{sec: labeling process}

The annotation process followed five steps. See \Cref{appendix: annotator roles and contributions} for additional details.

(1) Extraction: Student annotators, after receiving 5–10 hours of training, extracted relevant clauses for two contract categories from the Contract Corpus. For the remaining seven categories, we reused expert-annotated clauses from the CUAD dataset.

(2) Retrieval: Student annotators searched the Clause Corpus to retrieve relevant clauses for each query, aiming for 10 relevant (3- to 5-star) and 20 irrelevant (2-star) clauses per query. Some queries yielded only 5–8 relevant clauses.

(3) Scoring: Two experienced attorneys and one student annotator rated each query-clause pair using a detailed four-page rubric. Clauses mistakenly retrieved as relevant were assigned a 2-star rating.

(4) Reconciliation: For cases with rating discrepancies greater than 2 stars or disagreement on relevance, a panel of 3–6 experienced attorneys reviewed and adjusted scores to bring them within a 2-point range. Final scores were calculated as the average of the three individual ratings. 

(5) Expansion: To augment the dataset, 1-star irrelevant clauses were added to each query using clauses from the CUAD dataset.

\paragraph{Data Format} The ACORD dataset is released in BEIR format, with a modification to the format of the qrels files to account for our explicit 1-star ratings. See \Cref{appendix: data format} for details.

\section{Experiments}

\subsection{Setup}


\paragraph{Metrics} We use the following five standard information retrieval metrics to measure performance: NDCG@5, NDCG@10, 5-star precision@5, 4-star precision@5, 3-star precision@5.
NDCG measures the models’ ranking quality by how well they rank the most relevant clauses at the top of the search results, focusing on the top results.
The x-star-precision@5 metrics evaluate the precision within the top 5 results by counting how many of the top 5 results meet or exceed the respective relevance score. For NDCG, we change the scores from 1-5 to 0-4 to ensure the computed metrics correctly reflect the quality of returned results. See \Cref{appendix: performance metrics} for details and definitions of the performance metrics.

\paragraph{Baseline Models} We evaluate several stand-alone retriever methods, including BM25 \citep{robertson_probabilistic_2009}, MiniLM bi-encoder (66M parameters) \citep{wang_minilm_2020} and OpenAI text embedding \citep{openai}.
BM25 is a computationally inexpensive lexical or keyword-based retriever. Documents with terms from the query have higher scores; rarer terms in the overall corpus have higher weighting, so they contribute more to the document score.
MiniLM bi-encoder is an embedding model trained via knowledge distillation from BERT on various embedding datasets, including MSMARCO.

We evaluate two-phase retrieval-reranker systems by reranking the top $100$ BM25\footnote{We focus on BM25 for these experiments as it is one of the most common retrievers. Due to space constraints, we move results for the MiniLM bi-encoder into \Cref{appendix: model performance}.}
outputs with the MiniLM cross-encoder (22.7M parameters), GPT-4o-mini,\footnote{Using \texttt{gpt-4o-mini-2024-07-18}} GPT-4o,\footnote{Using \texttt{gpt-4o-2024-08-06}} Llama-3.2-1B (1.23B parameters), and Llama-3.2-3B (3.21B parameters) \citep{gpt4o,meta_llama} models.
 

\paragraph{Fine-tuning}
The MiniLM cross-encoder is fine-tuned on the training dataset.
We choose the learning rate and the number of updates via grid search on the validation NDCG@10 score.
See \Cref{appendix: fine_tuning} for full training details.

\subsection{Results}
We show the results of most tested search systems on the test dataset in \Cref{tab: results_main} with full results in \Cref{appendix: model performance}. Overall, a BM25 retriever paired with pointwise GPT-4o performs best, fine-tuned models perform better than models without fine-tuning, and embedding models perform better than lexical models. However, results vary significantly by query and model. Also, results using a pointwise instead of pairwise reranking are generally much better in our experiments; see \Cref{tab: pairwise and pointwise results} for a comparison. 

\begin{table*}[t]
    \scriptsize
    \centering
    \begin{tabular}{ll|ccccc}
        \toprule
         &  & \textbf{NDCG@5} & \textbf{NDCG@10} & \textbf{3-star prec@5} & \textbf{4-star prec@5} & \textbf{5-star prec@5} \\
        \textbf{Retriever} & \textbf{Reranker} &  &  &  &  &  \\
        \midrule 
         OpenAI embeddings (small) & None & 55.1 & 55.8 & 50.5 & 34.7 & 8.3 \\
        \midrule
        OpenAI embeddings (large) & None & 62.1 & 64.1 & 58.6 & 38.9 & 11.0 \\
        \midrule
        \multirow[t]{6}{*}{BM25} & None & 52.5 & 54.0 & 50.9 & 38.9 & 9.0 \\
         & Cross-Encoder-MiniLM & 59.3 & 60.9 & 60.0 & 43.5 & 6.2 \\
         & GPT4o & \textbf{76.9} & \textbf{79.7} & \textbf{81.1} & \textbf{60.0} & \underline{17.2} \\
         & GPT4o-mini & \underline{75.2} & \underline{78.2} & \underline{78.6} & \underline{58.2} & \textbf{18.6} \\
         & Llama-1B & 13.8 & 14.4 & 13.0 & 10.5 & 4.1 \\
         & Llama-3B & 62.6 & 65.3 & 63.9 & 48.1 & 9.7 \\
        \bottomrule
    \end{tabular}
    \caption{
    Summary of the performance in \% of selected retrieval models on the ACORD test dataset. The table reports NDCG@5, NDCG@10, and normalized precision@5 (for 3-star, 4-star, and 5-star clauses) aggregated across categories. We highlight the highest and second-highest metrics in bold and underlined, respectively. See \Cref{tab: detailed results all tested models} in \Cref{appendix: model performance aggregated} for results for all tested retriever methods. These results use a pointwise reranking as they are generally much better than the results using pairwise reranking in our experiments; See \Cref{tab: pairwise and pointwise results}.
    }
    \label{tab: results_main}
\end{table*}

\paragraph{Performance of Methods} 
In retrieval-only systems, the large OpenAI text embedding model performed best. BM25 or MiniLM bi-encoder with large LLM rerankers significantly outperformed other methods. MiniLM bi-encoder with GPT4o reranker achieved the highest results, with an NDCG@5 of 79.1\% and an NDCG@10 of 81.2\%, while BM25 with GPT4o reranker was a close second, outperforming all other methods by a noticeable margin. Smaller LLMs performed much worse than larger ones; Llama 3.2 showed a difference of 40\%-points in most metrics between the 1B and 3B models.

\begin{figure}[t]
    \centering
    \includegraphics[width=1.0\linewidth]{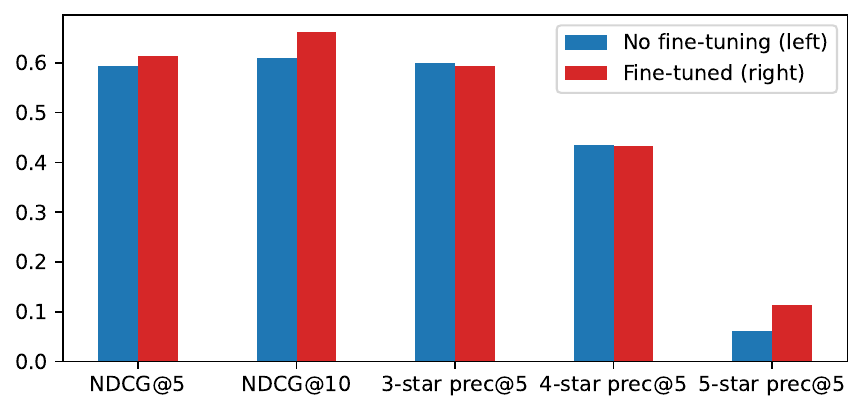}
    \caption{
        The performance of BM25 with the MiniLM cross-encoder before and after fine-tuning. We see improvements in NDCG and 5-star precision scores with minor declines in the 3-star and 4-star precision scores. These results underscore the importance of domain-specific fine-tuning for advancing clause retrieval capabilities; however, the improvements are not large, underlining that the problem is not solved. See \Cref{tab: cross-encoder finetuning detailed} in \Cref{appendix: model performance by category} for results for each category.
    }
    \label{fig: cross-encoder finetuning}
\end{figure}

\paragraph{Performance after Fine-Tuning} We finetuned the MiniLM cross-encoder on the training data and observed a modest increase of 2.0\% and 5.4\% in NDCG@5 and NDCG@10 scores, respectively. Meanwhile, the 3- and 4-star precisions decrease slightly, with a 5.1\% increase in 5-star precision. See \Cref{fig: cross-encoder finetuning} for results before and after fine-tuning, and \Cref{tab: cross-encoder finetuning detailed} in \Cref{appendix: model performance by category} for results by category. We also fine-tune the cross-encoder when paired with the bi-encoder for the initial retrievals; the results are in \Cref{tab: fine-tuned results}.

\paragraph{Performance by Clause Categories and Queries.} We measure the performance of the models across each of the 9 clause categories and under each query and find performance varies substantially. Some results are unexpected. For example, the query ``indemnity of broad-based claims'' achieved perfect scores for NDCG@5 and 3-star and 4-star precision@5 using BM25 paired with MiniLM cross-encoder, whereas ``unilateral liability cap'' and ``as-is clause'' got zero NDCG@5 score. This is contrary to what we observed from human annotators: extracting ``unilateral liability cap'' and ``as-is clauses'' was easier than extracting relevant ``indemnity of broad-based claims'' clauses. Results for each clause category can be found in \Cref{appendix: model performance by category}.

\section{Discussion and Future Work}\label{sec: discussion and future work}

\paragraph{Strong Extraction Performance} The models have demonstrated a strong understanding of legal terminology beyond simple lexical matching. For instance, in response to the query ``change of control'', the models effectively return clauses referencing ``ownership changes'' or the ``sale of substantially all assets''. Similarly, for the query ``IP ownership assignment/transfer,'' the models accurately identify ``work for hire'' clauses. The query ``indemnification of broad-based claims'' generates 100\% NDCG@5 and 3-star and 4-star precision@5 scores, showcasing deep semantic comprehension (broad-based claims mean claims brought by a contracting party or any third party for breach of contract and tort, etc.). This contributes to the overall promising NDCG@5 and 3-star precision@5 scores, indicating the models’ relative maturity in clause extraction tasks.

\paragraph{Poor Performance on Legal Jargon without Context} Despite the strong capability to understand legal terminologies, the models perform poorly on queries of legal jargon without context. For example, the query ``as-is clause'' achieves zero NDCG@5 score across several models, whereas a simple keyword search would yield near-perfect results. The query ``IP ownership assignment/transfer'' returns joint ownership clauses instead of straight-forward ownership assignment clauses. In contrast, models perform significantly better with longer, context-rich queries. The query ``IP infringement indemnity that covers trademark or copyright'' achieves near-perfect scores across several models. To further test the effectiveness of more contextualized queries, we expand two queries (``as-is clause'' and ``change of control'') into medium and long formats. We find that medium formats yield significantly better results, as shown in \Cref{appendix: query format results}. In the Supplemental Materials, we include expanded medium and long formats for each of the 114 queries. We invite the research community to experiment further using these formats as prompts.

\paragraph{Low Ranking Performance} Models consistently fail to retrieve 4- or 5-star clauses in the top 5 results, with the 4-star and 5-star precision@5 scores being only 60.0\% and 17.2\% for the overall best method (BM25 with GPT-4o pointwise reranking). Many 4- or 5-star clauses are ranked beyond the top 10 or even 15. Low-ranking performance diminishes user confidence and limits the broader adoption of these models for real-world contract-related tasks. This finding is not surprising, given that there is currently no expert-annotated IR dataset for contracts, and the gap highlights the need for large-scale expert-annotated IR datasets like ACORD.

\paragraph{NDCG Is Not an Optimal Metric} The high NDCG scores seen do not adequately reflect the quality of the models in real-life contract drafting tasks. In the context of clause retrieval, the more meaningful metrics are 4- and 5-star precision@5 scores. This is because ranking relevant clauses requires a high degree of legal expertise. Most users of legal IR applications may lack such expertise. Thus, presenting many 3-star clauses as high-quality results can mislead users into relying on poorly drafted precedents, ultimately resulting in low-quality contracts.

The best method, bi-encoder retriever with GPT4o pointwise reranking, had an NDCG@5 score of 79.1\%, but the 4-star and 5-star precision@5 scores were only 62.1\% and 17.2\%, respectively. Moreover, 40\% of the queries tested have a 4-star precision@5 score below 50\%. With ACORD, we hope the research community can advance in developing retrieval methodologies prioritizing clause quality. 

\paragraph{Pointwise Outperforms Pairwise} We test ACORD using pointwise and pairwise approaches on different LLMs with the results available in \Cref{tab: pairwise and pointwise results}. Interestingly, the pointwise approach performs better than the pairwise approach for all tested retrieval methods, except when using the smallest Llama 3.2 model as the reranker. The same result persists after we deploy several methodologies in our pairwise experiments, including Pairwise Ranking Prompting used by \citet{qin_large_2024}. This result seems inconsistent with prior literature, which generally indicates better results for pairwise methods \citep{qin_large_2024,zhuang_beyond_2024}. We encourage further research to validate and expand on our findings.

\section{Conclusion}

We introduced ACORD, the first expert-annotated clause retrieval dataset for contract drafting tasks. ACORD aims to facilitate research on clause retrieval, a foundational contract drafting task focusing on the most complex and heavily negotiated clauses in commercial contracts. ACORD includes 114 queries and over 126,000 query-clause pairs, each ranked on a scale from 1 to 5 stars. We evaluated 20 retrieval methods on ACORD and found that performance is promising but still has significant room for improvement.
We also showed that model design markedly influences performance, suggesting that algorithmic improvements from the NLP community will help solve this challenge. 

\section{Limitations}
ACORD is a simplified dataset because it contains a small scope of clauses already extracted from contracts. For example, ACORD does not have clauses about representatives and warranties, product descriptions, or payment terms. The performance will deteriorate in real-life legal tasks that would require the models to first extract relevant clauses from a large number of contracts before ranking them. We plan to publish the underlying contracts to enable research and experiments on the extraction and retrieval tasks. 

Expert annotator ratings largely depend on a lawyer’s individual experience, industry, and interpretation of the query, among other things. This subjectivity is reflected in the annotators' disagreement rate of 21\%. However, although the ratings may not be exact, the relative ranking among different query-clause pairs is less uncertain.  

The vast majority of commercial contracts are confidential and proprietary. ACORD uses publicly available sources, namely contracts in EDGAR and selected online terms of services. Although ACORD may not fully represent an enterprise's database of negotiated contracts, the types of queries and clauses it includes are fairly standardized, making ACORD a valuable resource for enterprise contract retrieval.

ACORD focuses primarily on English-language contracts for U.S.-based companies, which limits its applicability to contracts governed by other legal systems or written in other languages.

Expert lawyers generate queries in ACORD and likely do not reflect the queries that a less experienced non-legal professional would have asked. Translating these queries into plain English would require legal expertise. This highlights the importance of keeping human legal professionals in the loop when creating products using ACORD.

\section{Acknowledgements}

A full list of contributors to the ACORD dataset is available at \url{https://www.atticusprojectai.org/ACORD}.

\newpage

\bibliography{custom}
\bibliographystyle{acl_natbib}

\clearpage
\onecolumn
\appendix

\section{Appendix -- Annotation Details}\label{appendix: annotation rubrics}
\setcounter{table}{0}
\renewcommand{\thetable}{A-\arabic{table}}
\setcounter{figure}{0}
\renewcommand{\thefigure}{A-\arabic{figure}}

\begin{table*} [h]
    \centering
    \renewcommand{\arraystretch}{1.5}
    \scriptsize
    \begin{tabular}{l|m{2.5cm}|m{12.1cm}}
        \toprule
        \textbf{Rating} & \textbf{Label} & \textbf{Description} \\
        \midrule
        1 & Not relevant, not useful & The clause is not responsive to the query at all. \\
        
        2 & Not relevant, but useful & The clause is not responsive to the query but contains language that could be helpful in drafting. \\
        
        3 & Relevant, but not perfect & The clause is responsive to the query but has some defects. Such defects include, among other things, the clause: (1) is too long, (2) is missing key concepts typically seen in a real-life contract clause, (3) does not use standard legal language typically seen in a real-life contract, (4) is too one-sided and unfair, or (5) is unclear, etc. \\
        
        4 & Perfect & The clause is relevant, concise, clear, and covers all necessary elements of the query. \\
        
        5 & Exemplary & The clause meets all criteria for a 4 rating AND includes additional helpful legal concepts that may be responsive to the query. \\
        \bottomrule
    \end{tabular}
    \caption{
    ACORD Annotation rubric. Ratings range from 1 (not relevant) to 5 (exemplary), with each level defined by clear criteria. 
    }
    \label{tab: annotation rubrics}
\end{table*}

\subsection{Annotator Instructions and Interface}

\paragraph{Annotator Instructions}
Annotators were provided the five-point relevance rubric in \cref{tab: annotation rubrics} and the following instructions:

Please rate each clause assigned to you on a scale of 1-5 based on the following rubric. Clauses receiving a “Yes” for Relevance should receive a 3-5 rating. Clauses receiving a “No” for Relevance should receive a 1-2 rating. 

\paragraph{Annotation Interface}
Annotators complete Google Forms that include the query-clause pairs by rating each pair with a 3 to 5 star (or 2 if the clause is irrelevant). Annotators used Google Sheet to annotate the 2-star rated clauses. Results from completed Google Forms are then consolidated into the master dataset.

\subsection{Annotator Recruitment and Demographics}
A nonprofit organization recruited the annotators to create ACORD. The lawyer annotators are volunteers and do not receive compensation. Student annotators receive hourly compensation. All annotators were informed of the purpose of their involvement and consented to the inclusion of their annotations in ACORD.

Out of the annotators who consented to the disclosure of their demographic information, 42\% identify as male and 58\% identify as female. Except for one annotator based in the United Kingdom, all annotators are based in the United States.

\subsection{Annotator Roles and Contributions}\label{appendix: annotator roles and contributions}
Our annotation team consisted of 12 experienced attorneys and 10 students working under the attorneys' direct supervision following 5-10 hours of training. As detailed in \Cref{sec: labeling process} (``Labeling Process''), each clause pair in the dataset was independently annotated by three individuals: one student annotator and two experienced attorneys. In instances of disagreement, annotations were reviewed and adjudicated by a panel comprising three to six experienced lawyers to ensure consistency and legal accuracy. Annotator disagreement was tracked by the number of scores revised during this reconciliation step. 

A substantial proportion of 1-star clauses in the dataset originated from data augmentation using the CUAD dataset. 

All attorney contributions were provided on a pro bono basis. Student annotators received hourly compensation ranging from \$20 to \$35.

\newpage
\section{Appendix -- Query-Clause Pair with Score}\label{appendix: query-clause pair}
\setcounter{table}{0}
\renewcommand{\thetable}{B-\arabic{table}}
\setcounter{figure}{0}
\renewcommand{\thefigure}{B-\arabic{figure}}
Sample of an annotated query-clause pair.

\begin{table}[h]
    \scriptsize
    \centering
    \begin{tabular}{m{0.07\linewidth}|m{.8\linewidth}|m{0.04\linewidth}}
        \toprule
        \textbf{Query} & \textbf{Clause} & \textbf{Score} \\
        \midrule
        IP infringement exception to indirect damage waiver & 
        \textbf{8 INDEMNIFICATION.} \newline
        8.1 By Commerce One. Commerce One shall indemnify, defend and hold harmless Corio and its Customers from any and all damages, liabilities, costs and expenses (including reasonable attorneys' fees) incurred by Corio or its Customers arising out of any claim that the software infringes any patent, copyright, trademark or trade right secret of a third party; <omitted> \newline
        
        \textbf{9 LIMITATION OF LIABILITY.} \newline
        EXCEPT FOR LIABILITY ARISING UNDER SECTION 8 OF THIS AGREEMENT, IN NO EVENT SHALL EITHER PARTY’S LIABILITY ARISING OUT OF THIS AGREEMENT OR THE USE OR PERFORMANCE OF THE SOFTWARE EXCEED THE TOTAL AMOUNT ACTUALLY PAID BY CORIO HEREUNDER FOR THE TRANSACTION WHICH THE LIABILITY RELATES TO DURING THE TWELVE (12) MONTHS IMMEDIATELY PRIOR TO THE FILING OF THE CAUSE OF ACTION TO WHICH THE LIABILITY RELATES. EXCEPT FOR LIABILITY ARISING UNDER SECTION 8 OF THIS AGREEMENT, IN NO EVENT SHALL EITHER PARTY HAVE ANY LIABILITY TO THE OTHER PARTY FOR ANY LOST PROFITS OR COSTS OF PROCUREMENT OF SUBSTITUTE GOODS OR SERVICES, OR FOR ANY INDIRECT, SPECIAL OR CONSEQUENTIAL DAMAGES HOWEVER CAUSED AND UNDER ANY THEORY OF LIABILITY AND WHETHER OR NOT SUCH PARTY HAS BEEN ADVISED OF THE POSSIBILITY OF SUCH DAMAGE. THE PARTIES AGREE THAT THIS SECTION 9 REPRESENTS A REASONABLE ALLOCATION OF RISK. 
        & 4.67 \\
        \bottomrule
    \end{tabular}
    \caption{A sample query-clause pair with a score averaged from three expert annotations.}
    \label{tab: sample query-clause score}
\end{table}

\newpage
\section{Appendix -- Data Details}\label{appendix: data details}
\setcounter{table}{0}
\renewcommand{\thetable}{C-\arabic{table}}
\setcounter{figure}{0}
\renewcommand{\thefigure}{C-\arabic{figure}}

\subsection{Licensing}
ACORD is licensed under CC-BY-4.0.

\subsection{Ethics Board Review}
This research project was reviewed for ethics considerations by the ETH Zürich Ethics Commission.

\subsection{Data List and Statistics}\label{appendix: data list and statistics}
This subsection presents some details of the data. \Cref{tab: categories and queries} shows all 114 queries and the categories they belong to. While \Cref{tab: clause statistics,tab: ACORD queries and clauses counts} show summary statistics of the number of clauses for each query split by rating and counts for the number of clauses for each category and rating.
\renewcommand{\arraystretch}{1.2}
\begin{longtable}{l|p{3.6cm}|p{10.7cm}}
    \toprule
    \textbf{No.} & \textbf{Clause Category} & \textbf{Query} \\ \midrule
    \endfirsthead

    \toprule
    \textbf{No.} & \textbf{Clause Category} & \textbf{Query} \\
    \midrule
    \endhead
    
    \endfoot
    
    \endlastfoot  
    1 & Limitation of Liability & cap on liability \\ 
    2 & Limitation of Liability & liability cap is based on purchase price \\ 
    3 & Limitation of Liability & precedents for insurance coverage influencing limitation of liability \\ \
    4 & Limitation of Liability & Fix fee liability cap \\ 
    5 & Limitation of Liability & Cap on liability equals 12 months payment \\ 
    6 & Limitation of Liability & cap on liability for indirect damages \\ 
    7 & Limitation of Liability & unilateral liability cap \\ 
    8 & Limitation of Liability & mutual liability cap \\ 
    9 & Limitation of Liability & two parties having different liability caps and/or carveouts \\ 
    10 & Limitation of Liability & Cap on liability without carveouts \\ 
    11 & Limitation of Liability & liability cap carveouts \\ 
    12 & Limitation of Liability & cap on liability subject to law \\ 
    13 & Limitation of Liability & compliance with law carveout to cap on liability \\ 
    14 & Limitation of Liability & indemnification carveout to cap on liability \\ 
    15 & Limitation of Liability & IP infringement exception to cap on liability \\ 
    16 & Limitation of Liability & third party IP infringement exception to cap on liability \\ 
    17 & Limitation of Liability & liability cap clauses that exclude third party IP infringement and fraud, gross negligence or willful misconduct \\ 
    18 & Limitation of Liability & personal or bodily injury exception to liability cap \\ 
    19 & Limitation of Liability & confidentiality exceptions to liability cap \\ 
    20 & Limitation of Liability & fraud, negligence or willful misconduct carveout to liability cap \\ 
    21 & Limitation of Liability & a party’s liability for fraud, negligence, personal injury or tort subject to a cap \\ 
    22 & Limitation of Liability & personal or bodily injury exception to cap on liability via indemnification carveout \\ 
    23 & Limitation of Liability & third party IP infringement exception to cap on liability via indemnification carveout \\ 
    24 & Limitation of Liability & fraud, gross negligence or willful misconduct exception to cap on liability via indemnification carveout \\ 
    25 & Limitation of Liability & seller-favorable cap on liability clauses \\ 
    26 & Limitation of Liability & buyer-favorable cap on liability clauses \\ 
    27 & Limitation of Liability & non-reliance clause \\ 
    28 & Limitation of Liability & as-is clause \\ 
    29 & Limitation of Liability & as-is clause with carveouts \\ 
    30 & Limitation of Liability & unqualified ``as-is'' clause \\ 
    31 & Limitation of Liability & warranty disclaimer clause that includes implied warranties \\ 
    32 & Limitation of Liability & warranty disclaimer clause that disclaims implied warranties of merchantability and fitness for a particular purpose \\ 
    33 & Limitation of Liability & waiver of implied warranty of title and non-infringement \\ 
    34 & Limitation of Liability & warranty disclaimer that does not specifically waive title and non-infringement warranties \\ 
    35 & Limitation of Liability & exclusive remedy for breach of product warranty \\ 
    36 & Limitation of Liability & product replacement, repair or refund as exclusivity remedy \\ 
    37 & Limitation of Liability & product warranty of shelf life \\ 
    38 & Limitation of Liability & product warranty around manufacturing and shipping \\ 
    39 & Limitation of Liability & customer's right for defective products \\ 
    40 & Limitation of Liability & seller-favorable warranty disclaimer clauses \\ 
    41 & Limitation of Liability & buyer-favorable warranty disclaimer clauses \\  
    42 & Limitation of Liability & disclaimer of indirect damages \\ 
    43 & Limitation of Liability & consequential damages waiver \\ 
    44 & Limitation of Liability & incidental damages disclaimer \\ 
    45 & Limitation of Liability & disclaimer of lost profits \\  
    46 & Limitation of Liability & disclaimer of punitive damages \\ 
    47 & Limitation of Liability & disclaimer of strict liability \\ 
    48 & Limitation of Liability & unilateral indirect damages waiver \\ 
    49 & Limitation of Liability & mutual indirect damages waiver \\ 
    50 & Limitation of Liability & indirect damage waiver is subject to law \\ 
    51 & Limitation of Liability & indemnification carveout to indirect damage waiver \\ 
    52 & Limitation of Liability & IP infringement exception to indirect damage waiver \\  
    53 & Limitation of Liability & third party IP infringement exception to indirect damage waiver \\  
    54 & Limitation of Liability & indirect damage waiver clauses that exclude third party IP infringement and fraud, gross negligence or willful misconduct \\  
    55 & Limitation of Liability & personal or bodily injury exception to indirect damage waiver \\  
    56 & Limitation of Liability & confidentiality exceptions to indirect damage waiver \\ 
    57 & Limitation of Liability & fraud, gross negligence or willful misconduct carveout to indirect damage waiver \\ 
    58 & Limitation of Liability & indirect damages waiver applies to a party's liability for fraud, negligence or personal injury \\  
    59 & Limitation of Liability & personal or bodily injury exception to waiver of indirect damages via indemnification carveout \\  
    60 & Limitation of Liability & first party claim exception to waiver of indirect damages \\  
    61 & Limitation of Liability & fraud, gross negligence or willful misconduct exception to waiver of indirect damages via indemnification carveout \\  
    62 & Limitation of Liability & seller-favorable waiver of indirect damages clauses \\  
    63 & Limitation of Liability & buyer-favorable waiver of indirect damages clauses \\  
    64 & Indemnification & indemnity or indemnification clause \\  
    65 & Indemnification & mutual indemnification provisions \\  
    66 & Indemnification & unilateral indemnification clause \\  
    67 & Indemnification & indemnification of third party claims \\  
    68 & Indemnification & fraud and/or gross negligence indemnity \\  
    69 & Indemnification & third party IP infringement indemnity \\  
    70 & Indemnification & indemnification of third-party claims based on breach of agreement \\  
    71 & Indemnification & IP infringement indemnity that covers trademark or copyright \\  
    72 & Indemnification & indemnification covers indirect claims \\  
    73 & Indemnification & Indemnification of first party claims \\  
    74 & Indemnification & indemnification clauses that include hold harmless \\  
    75 & Indemnification & indemnification clauses that do not include hold harmless \\  
    76 & Indemnification & indemnification clauses that include defending claims \\  
    77 & Indemnification & indemnification clause that allows indemnifying party to control defenses \\  
    78 & Indemnification & Indemnity of broad-based claims \\  
    79 & Indemnification & Indemnified party includes affiliates \\  
    80 & Indemnification & first party indemnification of specified claims \\  
    81 & Indemnification & first party indemnification of broad-based claims \\  
    82 & Indemnification & indemnification of broad-based third party claims \\  
    83 & Indemnification & third party claim indemnity of fraud, negligence or willful misconduct \\  
    84 & Indemnification & Third Party claim indemnity limited to use of products in compliance with agreement \\  
    85 & Indemnification & indemnification that covers violation of law \\  
    86 & Indemnification & seller-favorable indemnification clauses \\  
    87 & Indemnification & buyer-favorable indemnification clauses \\  
    88 & Affirmative Covenants & Revenue/Profit Sharing \\  
    89 & Affirmative Covenants & Minimum Commitment \\  
    90 & Affirmative Covenants & Audit Rights \\  
    91 & Affirmative Covenants & Insurance \\  
    92 & Restrictive Covenants & Most Favored Nation Clause \\  
    93 & Restrictive Covenants & Non-compete bound by time and territory \\  
    94 & Restrictive Covenants & Exclusivity bound by time and territory \\  
    95 & Restrictive Covenants & No-Solicit of Customers \\  
    96 & Restrictive Covenants & No-Solicit Of Employees not bound by time or longer than 12 months \\  
    97 & Restrictive Covenants & Non-Disparagement \\  
    98 & Restrictive Covenants & Rofr/Rofo/Rofn \\  
    99 & Restrictive Covenants & Change Of Control \\  
    100 & Restrictive Covenants & Anti-Assignment clause that requires notice only for assignment to affiliates \\  
    101 & Restrictive Covenants & Covenant Not To Sue \\  
    102 & Term & Renewal clause that requires notice to Renew \\ 
    103 & Term & Clause that requires notice to terminate auto-renew \\ 
    104 & Term & Termination for Convenience \\ 
    105 & Governing Law & New York Governing Law \\ 
    106 & Governing Law & England Governing Law \\ 
    107 & Governing Law & Clause with multiple governing laws \\ 
    108 & Governing Law & Governing Law excluding UCC or other similar regulatory frameworks \\ 
    109 & Liquidated Damages & Liquidated Damages \\ 
    110 & Third party beneficiary & Third Party Beneficiary \\ 
    111 & IP ownership/license & IP Ownership Assignment or Transfer \\ 
    112 & IP ownership/license & Joint IP Ownership \\ 
    113 & IP ownership/license & License clause covering affiliates of licensor and/or licensee \\ 
    114 & IP ownership/license & Source Code Escrow \\ 
    \bottomrule
    \caption{The table lists the 9 clause categories and the 114 queries in ACORD.} \label{tab: categories and queries}
\end{longtable}

\vspace{1pt}
\begin{table}[h]
    \small
    \renewcommand{\arraystretch}{1.25}
    \centering
    \begin{tabular}{l|rrrrr}
        \toprule
        Score  & Min & Mean & Median & Std. Dev. & Max \\
        \midrule
        1-star & 395 & 1080.42 & 1293.00 & 383.28 & 1293 \\
        2-star & 15 & 20.13 & 20.00 & 2.50 & 28 \\
        3-star & 1 & 4.11 & 4.00 & 2.45 & 17 \\
        4-star & 1 & 5.82 & 6.00 & 2.48 & 12 \\
        5-star & 1 & 2.17 & 2.00 & 1.45 & 6 \\
        \bottomrule
    \end{tabular}
    \caption{The min, mean, median, standard deviation, and max statistics of the number of rated clauses for each score. Refer to \Cref{tab: ACORD queries and clauses counts} for the number of query-clause pairs in the data for each of the 9 clause categories.}
    \label{tab: clause statistics}
\end{table}

\begin{table}[h]
    \small
    \centering
    \begin{tabular}{l|r|rrrrr|r}
        \toprule
         & Number of Queries & \multicolumn{5}{c|}{Number of ratings in the dataset} & Total ratings \\
        Category &  & 5-star & 4-star & 3-star & 2-star & 1-star &  \\
        \midrule
        Limitation of Liability & 63 & 87 & 345 & 200 & 1277 & 81459 & 83368 \\
        Indemnification & 24 & 29 & 153 & 95 & 459 & 31032 & 31768 \\
        Affirmative Covenants & 4 & 4 & 27 & 17 & 80 & 1581 & 1709 \\
        Restrictive Covenants & 10 & 3 & 64 & 45 & 206 & 3956 & 4274 \\
        Term & 3 & 0 & 20 & 12 & 66 & 1187 & 1285 \\
        Governing Law & 4 & 2 & 19 & 10 & 82 & 1582 & 1695 \\
        Liquidated Damages & 1 & 0 & 6 & 4 & 21 & 395 & 426 \\
        Third-party beneficiary & 1 & 0 & 8 & 3 & 20 & 395 & 426 \\
        IP ownership/license & 4 & 5 & 21 & 17 & 84 & 1581 & 1708 \\
        \hline
        Total & 114 & 130 & 663 & 403 & 2295 & 123168 & 126659 \\
        \bottomrule
    \end{tabular}
    \caption{The table lists the number of queries and annotated clauses under each x-star precision@5 rating for each clause category in ACORD.}
    \label{tab: ACORD queries and clauses counts}
\end{table}

\subsection{Clause Extraction Methodology in Clause Corpus}\label{appendix: clause extraction methodology}

\begin{figure}[ht]
    \centering
    \includegraphics[width=0.8\linewidth]{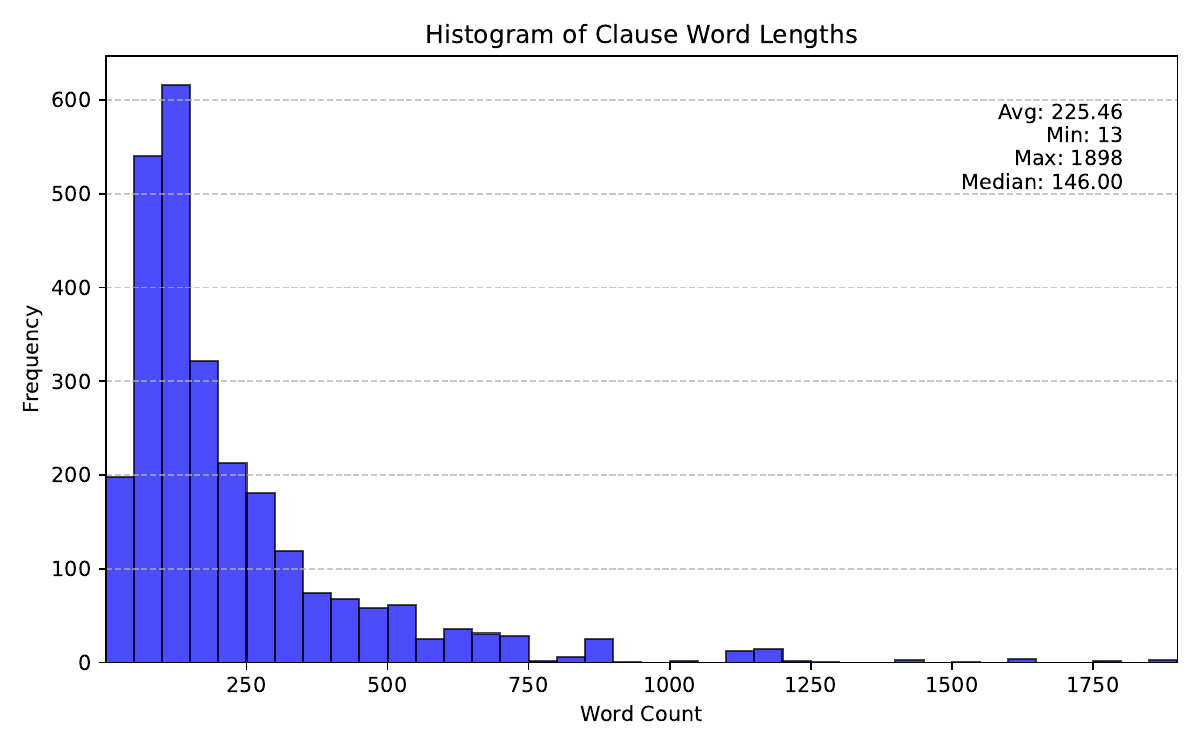}
    \caption{Histogram over the number of words in 2- to 5-star clauses in the ACORD dataset. The figure also includes statistics showing that 2- to 5-star clauses have an average of 225 words and a median of 146 words in each clause.}
    \label{fig: word count distribution}
\end{figure}

Most clauses are extracted at a section or subsection level, marked by a numerical header such as ``Section 9'' or ``8(a)''. If a single sentence within a section or subsection is relevant to the query, the entire section or subsection is included. Additionally, if the section or subsection references another section, such other section is also included. An <omitted> symbol is inserted between them to indicate that the two sections are not continuous in the contract.
Clauses have widely varying lengths, ranging from 13 words to 1,898 words. \Cref{fig: word count distribution} shows the distribution of 2- through 5-star clause lengths in ACORD.

\subsection{Data Format}\label{appendix: data format}

We release the ACORD dataset in BEIR format, with a modification to the format of the qrels files to account for explicit annotation of 1-star judgements.

\subsubsection{BEIR format}
In the BEIR format for ad-hoc IR datasets, the clauses and queries can be found in \texttt{corpus.jsonl} and \texttt{queries.jsonl}, respectively. The query-clause scores for the train, development, and test splits can be found in \texttt{qrels/{train, test, valid}.tsv}. As mentioned in the dataset section, lawyer judgments are one-indexed (between one and five stars), but it’s desirable to have zero-indexed scores when calculating NDCG. Therefore, in the qrels files, we subtract one from each lawyer's judgment to get scores that range from zero to four.

\subsubsection{Explicit zero-scores} We deviate from the standard BEIR format in that we explicitly, rather than implicitly, encode irrelevant query-clause pairs. In the standard BEIR format, irrelevant query-clause pairs (with a score of zero) are omitted from the qrels files, which implies their irrelevance. As with all Crenshaw- or TREC-style IR datasets where unjudged query-clause pairs are treated as irrelevant, this can lead to many false negatives (Thakur et al., 2021). In ACORD qrel files, we explicitly label irrelevant clauses with a score of zero. Omitted query-clause pairs should be ignored during evaluation because their score is unknown.

In constructing ACORD, we carefully constructed our dataset to reduce the incidence of false negatives. We explicitly label irrelevant clauses through lawyer judgments and through selecting clause sources that are known to be irrelevant to particular queries.


\newpage
\section{Appendix -- Model Performance}\label{appendix: model performance}
\setcounter{table}{0}
\renewcommand{\thetable}{D-\arabic{table}}
The next two sections will show detailed and extended results for the model evaluation of the test data.

\subsection{Model Performance by Category}\label{appendix: model performance by category}

\begin{table}[h]
    \scriptsize
    \centering
    \begin{tabular}{lrrrrr}
        \toprule
         & \textbf{NDCG@5} & \textbf{NDCG@10} & \textbf{3-star precision@5} & \textbf{4-star precision@5} & \textbf{5-star precision@5} \\
        \textbf{Category} &  &  &  &  &  \\
        \midrule       
        Limitation of Liability & 0.750 & 0.785 & 0.771 & 0.557 & 0.200 \\
        Indemnification & 0.763 & 0.803 & 0.800 & 0.600 & 0.178 \\
        Affirmative Covenants & 0.930 & 0.910 & 1.000 & 0.867 & 0.000 \\
        Restrictive Covenants & 0.682 & 0.662 & 0.850 & 0.650 & 0.200 \\
        Term & 0.912 & 0.936 & 1.000 & 0.600 & NaN \\
        Governing Law & 0.805 & 0.868 & 0.800 & 0.500 & 0.200 \\
        Liquidated Damages & 0.812 & 0.788 & 0.800 & 0.600 & NaN \\
        Third-party beneficiary & 1.000 & 0.974 & 1.000 & 1.000 & NaN \\
        IP ownership/license & 0.764 & 0.787 & 0.867 & 0.600 & 0.067 \\
        \bottomrule
    \end{tabular}
    \caption{The table shows the performance by clause category when using BM25 as the retriever model paired with pointwise GPT4o for reranking. We get Not a Number, NaN, for the 5-star precison@5 if there are no 5-star clauses for the queries in the category. Cells show the mean over test queries in a given category where we ignore NaN values.}
    \label{tab: BM25 GPT4o detailed}
\end{table}

In this section, we break the model performance down by category to determine how the different categories influence the performance of the models. We show in \Cref{tab: BM25 GPT4o detailed} the results for BM25 with GPT4o as the reranker model. Overall, we see that the 5-star precision@5 is struggling across all categories, but in particular, with ``Affirmative Covenants'' and ``IP ownership/license'', resulting in a low overall mean. However, for ``Affirmative Covenants,'' the NDCG scores are some of the highest across categories, thus showing that NDCG might not be the best metric in this setting as discussed in \Cref{sec: discussion and future work}.
\Cref{tab: cross-encoder finetuning detailed} shows the results by category before and after fine-tuning the cross-encoder on the training data. Overall, the improvement by fine-tuning the data is limited, which could indicate that more data or better architectures are needed. 

\begin{table}[h]
    \scriptsize
    \centering
    \begin{tabular}{lc|ccccc}
        \toprule
         &  & \textbf{NDCG@5 }& \textbf{NDCG@10} & \textbf{3-star precision@5} & \textbf{4-star precision@5} & \textbf{5-star precision@5} \\
        \textbf{Category} & \textbf{Fine-tuned} &  &  &  &  &  \\
        \midrule
        Limitation of Liability & \multirow[c]{9}{*}{No} & 0.541 & 0.565 & 0.507 & 0.364 & 0.043 \\
        Indemnification &  & 0.721 & 0.727 & 0.743 & 0.529 & 0.089 \\
        Affirmative Covenants &  & 0.605 & 0.571 & 0.667 & 0.533 & 0.000 \\
        Restrictive Covenants &  & 0.566 & 0.539 & 0.750 & 0.500 & 0.000 \\
        Term &  & 0.558 & 0.684 & 0.400 & 0.200 & NaN \\
        Governing Law &  & 0.517 & 0.544 & 0.600 & 0.500 & 0.200 \\
        Liquidated Damages &  & 0.838 & 0.765 & 1.000 & 0.600 & NaN \\
        Third-party beneficiary &  & 0.887 & 0.941 & 1.000 & 0.800 & NaN \\
        IP ownership/license &  & 0.382 & 0.458 & 0.333 & 0.333 & 0.067 \\
        \midrule
        Limitation of Liability & \multirow[c]{9}{*}{Yes} & 0.642 & 0.694 & 0.598 & 0.448 & 0.138 \\
        Indemnification &  & 0.678 & 0.735 & 0.638 & 0.500 & 0.081 \\
        Affirmative Covenants &  & 0.373 & 0.403 & 0.444 & 0.222 & 0.133 \\
        Restrictive Covenants &  & 0.565 & 0.509 & 0.717 & 0.567 & 0.067 \\
        Term &  & 0.447 & 0.583 & 0.200 & 0.133 & NaN \\
        Governing Law &  & 0.655 & 0.768 & 0.633 & 0.367 & 0.200 \\
        Liquidated Damages &  & 0.245 & 0.328 & 0.333 & 0.133 & NaN \\
        Third-party beneficiary &  & 0.841 & 0.859 & 0.933 & 0.533 & NaN \\
        IP ownership/license &  & 0.424 & 0.499 & 0.422 & 0.222 & 0.067 \\
        
        \bottomrule
    \end{tabular}
    \caption{The performance by clause category of the BM25 retriever model with the cross-encoder reranker model both with and without fine-tuning the cross-encoder on the training data. The finetuning hyperparameters were tuned by maximizing the mean NDCG@5 score for the validation set. We get Not a Number, NaN, for the 5-star precison@5 if there are no 5-star clauses for the queries in the category. Cells show the mean over all test queries in a given category where we ignore NaN values.}
    \label{tab: cross-encoder finetuning detailed}
\end{table}

\subsection{Model Performance and Size for All Tested Models}\label{appendix: model performance aggregated}
In this subsection, we show the results for all tested models using the test data. The results can be seen in \Cref{tab: detailed results all tested models} and show that the best and second best rerankers are GPT4o and GPT4o-mini when using either BM25 or Bi-Encoder-MiniLM. For the Llama-based rerankers, we see that the 3B size model far outperforms the 1B model. Thus, the model size has a large impact both when using BM25 and when using Bi-Encoder-MiniLM as the initial retriever. 

\begin{table}[t]
    \scriptsize
    \centering
    \begin{tabular}{ll|cccccr}
        \toprule
         &  & \textbf{NDCG@5} & \textbf{NDCG@10} & \textbf{3-star prec@5} & \textbf{4-star prec@5} & \textbf{5-star prec@5} & \textbf{Size}\\
        \textbf{Retriever} & \textbf{Reranker} &  &  &  &  &  & \\
        \midrule 
         OpenAI embeddings (small) & None & 0.551 & 0.558 & 0.505 & 0.347 & 0.083 & N/A \\
        \midrule
        OpenAI embeddings (large) & None & 0.621 & 0.641 & 0.586 & 0.389 & 0.110 & N/A \\
        \midrule
        \multirow[t]{6}{*}{BM25} & None & 0.525 & 0.540 & 0.509 & 0.389 & 0.090 & 0M \\
         & Cross-Encoder-MiniLM & 0.593 & 0.609 & 0.600 & 0.435 & 0.062 & 66M \\
         & GPT4o & \textbf{0.769} & \textbf{0.797} & \textbf{0.811} & \textbf{0.600} & \underline{0.172} & 1.8T\\
         & GPT4o-mini & \underline{0.752} & \underline{0.782} & \underline{0.786} & \underline{0.582} & \textbf{0.186} & 8B\\
         & Llama-1B & 0.138 & 0.144 & 0.130 & 0.105 & 0.041 & 1.23B \\
         & Llama-3B & 0.626 & 0.653 & 0.639 & 0.481 & 0.097 & 3.21B \\
        \midrule
        \multirow[t]{6}{*}{Bi-Encoder-MiniLM} & None & 0.571 & 0.572 & 0.498 & 0.358 & 0.076 & 66M \\
         & Cross-Encoder-MiniLM & 0.601 & 0.610 & 0.586 & 0.428 & 0.069 & 132M \\
         & GPT4o & \textbf{0.791} & \textbf{0.812} & \textbf{0.814} & \textbf{0.621} & \underline{0.172} & 1.8T \\
         & GPT4o-mini & \underline{0.763} & \underline{0.790} & \underline{0.775} & \underline{0.586} & \textbf{0.200} & 8B \\
         & Llama-1B & 0.200 & 0.205 & 0.147 & 0.119 & 0.048 & 1.29B \\
         & Llama-3B & 0.628 & 0.657 & 0.628 & 0.481 & 0.110 & 3.27B \\
        \bottomrule
    \end{tabular}
    \caption{The table shows the results for all tested models on the test data. For each of the retrieval models BM25 and Bi-Encoder-MiniLM, we bold and underline the best and second best scores, respectively. We also include the size of the models. For GPT4o and GPT4o-mini, the exact sizes are unknown when writing, but we use estimates found online \citet{GPT4o_size_estimate}. For the OpenAI embedding models, the exact sizes are unknown, and we did not find estimates.}
    \label{tab: detailed results all tested models}
\end{table}

\subsection{Model Performance after Fine-tuning}
We show below in \Cref{tab: fine-tuned results} the results of fine-tuning the cross-encoder reranker to the training data before evaluating the methods on the test data. We tested with both BM25 and bi+encoder as the initial retriever models. Overall, fine-tuning improves the models slightly; however, the results are still worse than using GPT4o or GPT4o-mini as the rerankers, cf. \Cref{tab: detailed results all tested models}. 

\begin{table*}[h]
    \scriptsize
    \centering
    \begin{tabular}{ll|ccccc}
        \toprule
         &  & \textbf{NDCG@5} & \textbf{NDCG@10} & \textbf{3-star prec@5} & \textbf{4-star prec@5} & \textbf{5-star prec@5} \\
        \textbf{Retriever} & \textbf{Reranker} &  &  &  &  &  \\
        \midrule 
         \multicolumn{7}{c}{Not fine-tuned} \\
         \midrule
        BM25 & Cross-Encoder-MiniLM & 59.3 & 60.9 & 60.0 & 43.5 & 6.2 \\
        Bi-Encoder-MiniLM & Cross-Encoder-MiniLM & 60.1 & 61.0 & 58.6 & 42.8 & 6.9 \\
         \midrule
         \multicolumn{7}{c}{Fine-tuned} \\
         \midrule
        BM25 & Cross-Encoder-MiniLM & 61.3 (+2.0) & 66.3 (+5.4) & 59.4 (-0.6) & 43.3 (-0.2) & 11.3 (+5.1) \\
        Bi-Encoder-MiniLM & Cross-Encoder-MiniLM & 64.7 (+4.6) & 68.1 (+7.1) & 61.6 (+3.0) & 44.9 (+2.1) & 12.2 (+5.3) \\
        \bottomrule
    \end{tabular}
    \caption{
        Performance in \% before and after fine-tuning the cross-encoder with pointwise reranking. We write the change in each metric after fine-tuning in parenthesis. We test both with BM25 and bi-encoder as the retriever model. Overall, the results show a modest improvement in NDCG scores and 5-star precision, as well as 3- and 4-star precisions for the bi-encoder model. However, with BM25, we see a slight decline in 3- and 4-star precisions. 
    }
    \label{tab: fine-tuned results}
\end{table*}

\subsection{Model Performance for Pairwise and Pointwise Rerankings}
We show below in \Cref{tab: pairwise and pointwise results} the results when using pairwise reranking rather than pointwise reranking. We tested it with GPT4o, GPT4o-mini, Llama 3.2 1B, and Llama 3.2 3B. Overall, we see that the results are much worse when using pairwise reranking. However, Llama 3.2 1B does see a significant improvement in its metrics, but it is still worse than the other models when they use pointwise reranking. 

\begin{table*}[h]
    \small
    \centering
    \begin{tabular}{ll|ccccc}
        \toprule
         &  & \textbf{NDCG@5} & \textbf{NDCG@10} & \textbf{3-star prec@5} & \textbf{4-star prec@5} & \textbf{5-star prec@5} \\
        \textbf{Retriever} & \textbf{Reranker} &  &  &  &  &  \\
        \midrule 
         \multicolumn{7}{c}{Pointwise} \\
         \midrule
        \multirow[t]{4}{*}{BM25} & GPT4o & 76.9 & 79.7 & 81.1 & 60.0 & 17.2 \\
         & GPT4o-mini & 75.2 & 78.2 & 78.6 & 58.2 & 18.6 \\
         & Llama-1B & 13.8 & 14.4 & 13.0 & 10.5 & 4.1 \\
         & Llama-3B & 62.6 & 65.3 & 63.9 & 48.1 & 9.7 \\
         \midrule
         \multicolumn{7}{c}{Pairwise} \\
         \midrule
        \multirow[t]{4}{*}{BM25} & GPT4o & 58.0 & 59.1 & 56.1 & 43.9 & 13.1 \\
         & GPT4o-mini & 57.4 & 58.5 & 55.4 & 43.5 & 12.4 \\
         & Llama-1B & 52.1 & 54.3 & 50.2 & 37.9 & 9.0 \\
         & Llama-3B & 52.8 & 54.4 & 50.5 & 38.9 & 9.0 \\
        \bottomrule
    \end{tabular}
    \caption{
        Performance in \% of GPT4o, GPT4o-mini, Llama 3.2 1B, and Llama 3.2 3B reranker when using either pointwise or pairwise reranking with BM25 as the retriever model. Overall, the pairwise results are much worse than those for pointwise reranking. However, Llama 3.2 1B does see a significant benefit from the more computationally heavy technique.
    }
    \label{tab: pairwise and pointwise results}
\end{table*}

\newpage
\section{Appendix -- Modifying Query Format}\label{appendix: query format results}
\setcounter{table}{0}
\renewcommand{\thetable}{E-\arabic{table}}
In this section, we present results for modifying the query to use a medium and long format rather than the default short queries. An expert lawyer constructs the medium and long-format queries. We show in \Cref{tab: query_variations sample} the results using different formats, demonstrating that providing more context to the models significantly improve performance. Although we only tested the ``Change of Control'' and ``as-is'' queries, in the Supplemental Materials, we provide medium and long format queries for all 114 queries and encourage the research community to conduct further experiments to improve the models' performance.

\begin{table}[H]
    \small
    \centering
    \begin{tabular}{c|c|p{0.6\textwidth}}
        \toprule
        \textbf{Query Type} & \textbf{Variation} & \textbf{Query} \\
        \midrule
        \multirow[c]{4}{*}{Change of control} 
        & Short & ``Change Of Control'' \\
        & Medium & ``Clause that prohibits change of control itself'' \\
        & Long & ``Does one party have the right to terminate or is consent or notice required of the counterparty if such party undergoes a change of control, such as a merger, stock sale, transfer of all or substantially all of its assets or business, or assignment by operation of law?'' \\
        \midrule
        \multirow[c]{4}{*}{'as-is'}
        & Short & ``'as-is' clause'' \\
        & Medium & ``'as-is' clause'' that disclaims all warranties' \\
        & Long & ``Is this a clause that states that the goods or services are being provided on an 'as-is' basis in their current condition, with no warranties or guarantees regarding their quality, performance, or suitability?'' \\
        \bottomrule
    \end{tabular}
    \caption{Query variations for two query types: ``Change of control'' and ``as-is''. Each variation represents a different length or level of detail, with queries ranging from short, concise statements to long, detailed descriptions.}
    \label{tab: query_variations sample}
\end{table}

\begin{table}[H]
\scriptsize
    \centering
    \begin{tabular}{lll|ccccc}
        \toprule
         &  &  & \textbf{NDCG@5} & \textbf{NDCG@10} & \textbf{3-star precision@5} & \textbf{4-star precision@5} & \textbf{5-star precision@5} \\
        \textbf{Model} & Query Type & Variation &  &  &  &  &  \\
        \midrule
        \multirow[t]{6}{*}{BM25} & \multirow[t]{3}{*}{Change of control} & Short & 0.914 & 0.856 & 0.800 & 0.600 & 0.200 \\
         &  & Medium & 0.887 & 0.794 & 1.000 & 0.600 & 0.200 \\
         &  & Long & 0.470 & 0.470 & 0.200 & 0.200 & 0.200 \\
        \cline{2-8}
         & \multirow[t]{3}{*}{'as-is'} & Short & 0.000 & 0.000 & 0.000 & 0.000 & 0.000 \\
         &  & Medium & 0.683 & 0.660 & 0.400 & 0.400 & 0.200 \\
         &  & Long & 0.284 & 0.297 & 0.200 & 0.200 & 0.000 \\
        \midrule
        \multirow[t]{6}{*}{BM25 + GPT-4o} & \multirow[t]{3}{*}{Change of control} & Short & 0.775 & 0.821 & 1.000 & 0.400 & 0.200 \\
         &  & Medium & 0.449 & 0.561 & 0.600 & 0.000 & 0.000 \\
         &  & Long & 0.835 & 0.850 & 1.000 & 0.200 & 0.200 \\
        \cline{2-8}
         & \multirow[t]{3}{*}{'as-is'} & Short & 0.102 & 0.068 & 0.000 & 0.000 & 0.000 \\
         &  & Medium & 1.000 & 0.954 & 1.000 & 1.000 & 0.000 \\
         &  & Long & 1.000 & 0.917 & 1.000 & 0.800 & 0.000 \\
         \midrule
        \multirow[t]{6}{*}{BM25 + Cross-Encoder} & \multirow[t]{3}{*}{Change of control} & Short & 0.832 & 0.770 & 0.800 & 0.600 & 0.200 \\
         &  & Medium & 0.590 & 0.707 & 0.400 & 0.200 & 0.000 \\
         &  & Long & 0.629 & 0.717 & 0.400 & 0.400 & 0.000 \\
        \cline{2-8}
         & \multirow[t]{3}{*}{'as-is'} & Short & 0.000 & 0.000 & 0.000 & 0.000 & 0.000 \\
         &  & Medium & 0.887 & 0.794 & 1.000 & 0.600 & 0.200 \\
         &  & Long & 0.518 & 0.641 & 0.600 & 0.200 & 0.000 \\
        \bottomrule
    \end{tabular}
    \caption{Comparison of retrieval performance across different query variations, categories, and models. The table reports five quality metrics (NDCG@5, NDCG@10, 3-star precision@5, 4-star precision@5, and 5-star precision@5) for each model (BM25, BM25 + GPT-4o, and BM25 + Cross-Encoder) evaluated on three different formats for the ``Change of Control'' and ``as-is'' queries. Using the more detailed queries, the medium or long formats, shown in \Cref{tab: query_variations sample}, give better results. However, the long format is not always an improvement over the medium format.}
    \label{tab: query formats results}
\end{table}

\section{Appendix -- Experiment and Evaluation Details}

\subsection{Performance Metrics}\label{appendix: performance metrics}
Performance is measured using standard metrics like NDCG@5, NDCG@10, 5-star-precision@5, 4-star-precision@5, and 3-star-precision@5. 
When computing the metrics, the scores are labeled from 0 to 4 rather than 1 to 5 to ensure irrelevant results (1-star clauses) are weighted appropriately.

The NDCG@k measures the normalized discounted cumulative gain of the top k returned results. We use the standard definition of NDCG@k seen in \eqref{eq: NDCG@k}.
\begin{align}
    DCG@k &= \sum_{i=1}^k \frac{rel_i}{\log_2(i+1)},\nonumber\\
    IDCG@k &= \sum_{i=1}^{|REL_k|} \frac{rel_i}{\log_2(i+1)},\nonumber\\
    NDCG@k &= \frac{DCG@k}{IDCG@k},\label{eq: NDCG@k}
\end{align}
where $rel_i$ is the graded relevance of the document at position $i$, and $REL_k$ is the list of relevant documents sorted by relevance. We use the \texttt{pytrec\_eval} library to calculate the NDCG scores.

The k-star precision@5 precision metric counts the number of returned clauses in the top 5 results rated $\geq k$. If there are less than 5 viable clauses for a query, i.e., clauses with ratings $\geq k$, then we normalize by the number of viable clauses. This ensures the precision metrics always range from 0 to 1. Formally, the k-star precision@5 is defined in \eqref{eq: precision at 5}.

\begin{align}
    k\text{-star precision}@5 &= \frac{\sum_{i=1}^5 \mathbbm{1}(rel_i\geq k)}{\min\left(5,\sum_i\mathbbm{1}(rel_i\geq k)\right)}\label{eq: precision at 5},
\end{align}
where $\mathbbm{1}$ is the indicator function so $\mathbbm{1}(rel_i\geq k)$ is $1$ if $rel_i\geq k$ and 0 otherwise.

\subsection{Fine-tuning Experiments}
\label{appendix: fine_tuning}

We perform a grid search over the learning rate in
$\{5 \times 10^{-5}, 1 \times 10^{-4}, 3 \times 10^{-4}, 5 \times 10^{-4}, 1 \times 10^{-3}\}$
and number of updates in $\{100, 200, 400, 600, 800, 1000, 1500, 2000, 2500, 3000\}$, choosing the hyperparameters with the best average validation NDCG@10 score over three runs.

We oversampled the data so that an equal number of examples of each relevance appear in the training data. We use a batch size of $64$ and the AdamW optimizer with a weight decay of $0.01$. We use cross-entropy loss, where the probability targets are the normalized relevance scores. For example, a one-star clause corresponds to a target of $p=0.0$, and a four-star clause corresponds to a target of $p=0.75$.

\subsection{Compute}
Fine-tuning experiments and Llama evaluations were performed on an NVIDIA RTX A6000 GPU. For fine-tuning experiments the full grid-search took about three GPU hours. Evaluation of bi-encoder and cross-encoder MiniLM models took less than five minutes per run. Llama pointwise evaluations took about thirty minutes per model, and pairwise evaluations took about an hour per model.

\end{document}